\documentclass[10pt,letterpaper]{article} 

\usepackage{iccv}
\usepackage{times}
\usepackage{epsfig}
\usepackage{graphicx}
\usepackage{amsmath}
\usepackage{amssymb}
\usepackage{subfigure}
\usepackage{tabulary}
\usepackage[pagebackref=true,breaklinks=true,letterpaper=true,colorlinks,bookmarks=false]{hyperref}
\usepackage[section]{placeins} 

\newcommand{\tld}{\raise.17ex\hbox{$\scriptstyle\mathtt{\sim}$}} 

\iccvfinalcopy 

\begin{document}

\title{FireCaffe: near-linear acceleration of deep neural network training \\ on compute clusters}

\author{Forrest N. Iandola, Khalid Ashraf, Matthew W. Moskewicz, Kurt Keutzer\\
\href{http://deepscale.ai}{DeepScale}\thanks{\href{http://deepscale.ai}{http://deepscale.ai}} ~ and UC Berkeley\\
{\tt\small \{forresti, ashrafkhalid, moskewcz, keutzer\}@eecs.berkeley.edu }
}

\maketitle
\vspace{-0.1in}
\begin{abstract}
Long training times for high-accuracy deep neural networks (DNNs) impede research into new DNN architectures and slow the development of high-accuracy DNNs. 
In this paper we present FireCaffe, which successfully scales deep neural network training across a cluster of GPUs.
We also present a number of best practices to aid in comparing advancements in methods for scaling and accelerating the training of deep neural networks. 	 
The speed and scalability of distributed algorithms is almost always limited by the overhead of communicating between servers; DNN training is not an exception to this rule.
Therefore, the key consideration here is to reduce communication overhead wherever possible, while not degrading the accuracy of the DNN models that we train.
Our approach has three key pillars.
First, we select network hardware that achieves high bandwidth between GPU servers -- Infiniband or Cray interconnects are ideal for this.
Second, we consider a number of communication algorithms, and we find that reduction trees are more efficient and scalable than the traditional parameter server approach. 
Third, we optionally increase the batch size to reduce the total quantity of communication during DNN training, and we identify hyperparameters that allow us to reproduce the small-batch accuracy while training with large batch sizes.
When training GoogLeNet and Network-in-Network on ImageNet, we achieve a 47x and 39x speedup, respectively, when training on a cluster of 128 GPUs.

\end{abstract}
\vspace{-0.1in}

\section{Introduction and Motivation}
\label{sec:intro}
\vspace{-0.1in}

Since the publication of AlexNet~\cite{alexnet}, a variety of new deep neural network (DNN) architectures such as GoogleNet~\cite{googlenet}, Network-in-Network~\cite{NiN}, and VGG~\cite{VGG-19} have been developed at a rapid pace. 
This is natural, because with the training and testing dataset fixed (e.g. ImageNet-1k~\cite{imagenet}), it is the DNN architecture that is primarily responsible for improvements in accuracy. 
In other words, the race for improvements in accuracy in image classification and other contemporary problems of computer science, has become a race in the development of new DNN architectures. 
So, what is the bottleneck in the development of new architectures?

In the development of new DNN architectures, as in any human research endeavor, creativity is a key element. 
However, the impact of architectural variations in DNNs -- such as number of layers, filter dimensions, and so forth -- can be hard to predict, and experimentation is required to assess their impact.
A high-accuracy deep neural network (DNN) model such as GoogLeNet~\cite{googlenet} can take weeks to train on a modern GPU. 
This is true even when leveraging deep neural network primitives like cuDNN~\cite{cuDNN}, maxDNN~\cite{maxDNN}, or fbfft~\cite{fbfft} -- all of which operate near the theoretical peak computation per second achievable on GPUs.  
Thus, training time is a key challenge at the root of the development of new DNN architectures. 
This sentiment was voiced by Jeffrey Dean of Google in his recent keynote address~\cite{DeanCIKM}. 

The four key points that Dean makes are: 
\begin{itemize}
	\item We [i.e. DNN researchers and users] want results of experiments quickly 
	\item There is a ``patience threshold:" No one wants to wait more than a few days or a week for a result
	\item This significantly affects scale of problems that can be tackled 
	\item We sometimes optimize for experiment turnaround time, rather than absolute minimal system resources for performing the experiment 
\end{itemize}

Given the considerable resources available to Google researchers, Dean's comments indicate that simply throwing more computational resources at the problem is not sufficient to solve the DNN training problem.  
In the following, we will spend a little more time dimensionalizing the current problems with DNN training and the upside potential if these problems can be solved. 

\subsection{Accelerating DNN Research and Development}

As a particular example of where long training times are limiting the pace of DNN research and productization, consider the following. 
ImageNet-1k has 1.2 million training images, distributed across 1000 different category labels. 
From first-hand conversations with engineers and executives, we know that several internet companies have internal databases containing billions of images with hundreds of thousands of different category labels. 
Due to long training times, these companies are facing serious delays in bringing DNN-based solutions to market. 
Accelerated DNN training solutions would solve a major pain point for these companies.

\subsection{Real-Time DNN Training}
So far, we have argued how accelerating DNN training would benefit applications where DNNs are in use today.
Now, we consider ways in which accelerating DNN training would allow DNN-based techniques to be applied in entirely new ways. 
There are a number of situations where it is crucial to incorporate new data into a DNN model in real time. 
For example, reinforcement learning (RL) enables robots to learn things themselves with minimal supervision. 
A recent study by Levine \etal applied state-of-the-art DNN-based RL techniques to enable a robot to teach itself how to build lego structures and screw on bottle caps~\cite{RL2015}. 
This technique is effective, and the robot does indeed learn to screw on bottle caps. 
However, it takes 3-4 hours for the robot to learn to screw on bottle caps, and the majority of this time is spent on DNN training. 
Faster DNN training would enable this and other reinforcement learning applications to move toward real-time.

Deep Neural Networks are used for an ever-broadening variety of problems, including classifying~\cite{googlenet, DeepLogo} and detecting~\cite{DenseNet, DPMareCNN} objects in images, writing sentences about images~\cite{forrestMicrosoft}, identifying actions in videos~\cite{ashraf15}, performing speech recognition~\cite{DeepSpeech}, and gaining semantic understanding of text~\cite{word2vec}.
We anticipate that sophisticated reinforcement learning (RL) systems in robotics will eventually leverage all of these modalities, ideally in real-time.

\subsection{Accelerating DNN Training with FireCaffe}
In our work, we focus directly on the problem of DNN training. 
Since single-GPU efficiency has reached the hard limits of the hardware, the next frontier for accelerating DNN training is to scale it across a compute cluster. 
In this paper, we present FireCaffe, which scales DNN training across a cluster of 128 GPUs with speedups of more than 40x compared to a single GPU. 
Our strategy for scaling up DNN training is to focus on reducing communication overhead, and we make a number of design choices toward this goal.
For example, we use fast interconnects such as Infiniband or Cray Gemini to accelerate communication among the GPUs.
We also show that reduction trees are a faster method for communication than using parameter servers. 
We map our parallelization strategy to high-accuracy DNN architectures that require less communication.

The rest of the paper is organized as follows.
In Section~\ref{sec:hardware}, we describe our choice of hardware for evaluating scalable DNN training, and Section~\ref{sec:preliminaries} introduces key factors that we will use for analyzing communication among GPU workers.
We describe tradeoffs between DNN parallelism strategies in Section~\ref{sec:parallelism-strategies}, and Section~\ref{sec:choosing-architectures} explains why certain high-accuracy DNN architectures are particularly amenable to parallelism.
In Section~\ref{sec:implementation}, we describe our approach to efficiently implementing distributed DNN training.
In Section~\ref{sec:eval}, we describe good practices that facilitate the comparison of scalable DNN training techniques and we present our speedups for training the NiN and GoogLeNet architectures on ImageNet.
Section~\ref{sec:related} describes approaches that are complimentary to FireCaffe for further accelerating DNN training.
Finally, we conclude in Section~\ref{sec:conclusions}.

\section{Hardware for scalable DNN training}
\label{sec:hardware}
\vspace{-0.1in}

It is both useful and possible to experiment with the scalability of DNN computations using theoretical or scale models.
However, demonstration and verification of the correctness and real-world scalability of the proposed FireCaffe system requires using concrete hardware platforms.
The speed at which data can be sent between nodes is a key consideration in selecting a hardware platform for scalable DNN training. 
This is because, the faster the interconnect between nodes is, the more scale we can achieve without being dominated by communication overhead.
Hardware manufacturers such as Cray and Mellanox address this by developing high-bandwidth, low-latency interconnects that are substantially faster than typical Ethernet connections.

For example, the Titan supercomputer at Oak Ridge Leadership Computing Facility (OLCF) has a high bandwidth, low latency Cray Gemini interconnect for communication among servers.
The Titan supercomputer has a total of 18000 servers, with one NVIDIA Kepler-based K20x GPU per server~\cite{ExascaleOLCF, IntroTitan}.
With this in mind, we choose the OLCF Titan supercomputer for tuning and evaluating FireCaffe.


In this research, we use relatively small slices of the overall capacity of Titan for each training run.
The additional computational capacity (\tld27 PetaFLOPS/s in total) enables us to conduct multiple training runs concurrently, where each training run utilizes 32 to 128 GPUs.
When considering 32-node slices of Titan, we found that the interconnect speed (at least for the applications of this work) is similar to that provided by having all nodes in the slice connected to a single Infiniband-class switch.


\section{Preliminaries and terminology}
\label{sec:preliminaries}
\vspace{-0.1in}
Deep neural network training is comprised of iterating between two phases: forward and backward propagation.
In the forward phase, a batch of data items (e.g. images) is taken from the training set, and the DNN attempts to classify them.
Then comes the backward phase, which consists of computing gradients with respect to the weights ($\nabla W$) and gradients with respect to the data ($\nabla D$).
The weight gradients are used to update the model's weights.
Then, an other forward phase is performed, and so on.
We train models using batched stochastic gradient descent (SGD), which is the standard choice for popular DNN models such as GoogLeNet~\cite{googlenet}.

We now present a few preliminaries, which we will use later in the paper for reasoning about data volume to communicate in distributed DNN training.
In Equation~\ref{eq:weightsize}, we show how to calculate the total size (in bytes) of the weights in all convolutional and fully-connected layers, combined.
\begin{equation} 
	\label{eq:weightsize}
     |W| = \sum_{L=1}^{\#layers} ch_L * numFilt_L * filterW_L * filterH_L * 4
\end{equation}
where $ch$ is the number of channels, $numFilt$ is the number of filters, $filterH$ is the filter height, and $filterW$ is the filter width. 
Next, Equation~\ref{eq:activationsize} expresses the size of activations produced by all layers, combined.
\begin{footnotesize}
\begin{equation} 
	\label{eq:activationsize}
	|D| = \sum_{L=1}^{\#layers} ch_L * numFilt_L * activationW_L * activationH_L * batch * 4
\end{equation}
\end{footnotesize}
where $activationH$ is the activation map height, $activationW$ is the activation width, and $batch$ is the batch size.
Note the $*4$ in Equations~\ref{eq:weightsize} and~\ref{eq:activationsize} -- this is because a floating-point number is 4 bytes.

To minimize confusion, we now define some terminology.
In our terminology, the each following sets of words are synonyms: (weights = parameters = filters = $W$); (nodes = workers = GPU servers).
We also sometimes use the terms ``activations" and ``data" ($D$) interchangeably.
{\em Fully-connected layers} are a special case of convolutional layers where $filterH=activationH$ and $filterW=activationW$.
We define an ``epoch" as one pass through the training data.
Finally, the word ``performance" can be ambiguous, so we write in terms of specific metrics such as ``accuracy" and ``training time."

\section{Parallelism strategies}
\label{sec:parallelism-strategies}
\vspace{-0.1in}
There are two commonly-used methods for parallelizing neural network training across multiple servers: model parallelism (e.g.~\cite{SpertII}) and data parallelism (e.g.~\cite{tencent}).

For batched SGD training of DNNs, we define {\em data parallelism} as the case where each worker (e.g. GPU) gets a subset of the batch, and then the workers communicate by exchanging weight gradient updates $\nabla W$.
We define {\em model parallelism} as the case where each worker gets a subset of the model parameters, and the workers communicate by exchanging data gradients $\nabla D$ and exchanging activations $D$.
Note that $|W| = |\nabla W|$ and $|D| = |\nabla D|$; in other words, the weights and weight gradients are the same size; and the data and data gradients are the same size.

Now, to maximize DNN training scalability, our goal is to select a parallelism strategy that requires the lowest possible quantity of communication between servers. 
The choice of whether it is ideal to use data parallelism, model parallelism, or both depends strongly on the DNN's architectural characteristics.
Commonly-used DNN architectures for speech recognition (e.g.~\cite{Seide11}) consist primarily of fully-connected layers, where the activations and parameters have the same spatial resolution (typically 1x1).
For typical batch sizes, these fully-connected models often have a similar quantity of weights $W$ and activations $D$.
For example, we observe in Table~\ref{T:data-volumes} that this property holds true for the MSFT-Speech DNN architecture~\cite{Seide11}.

\begin{figure}[!t]
	\centering
		\includegraphics[width=6in]{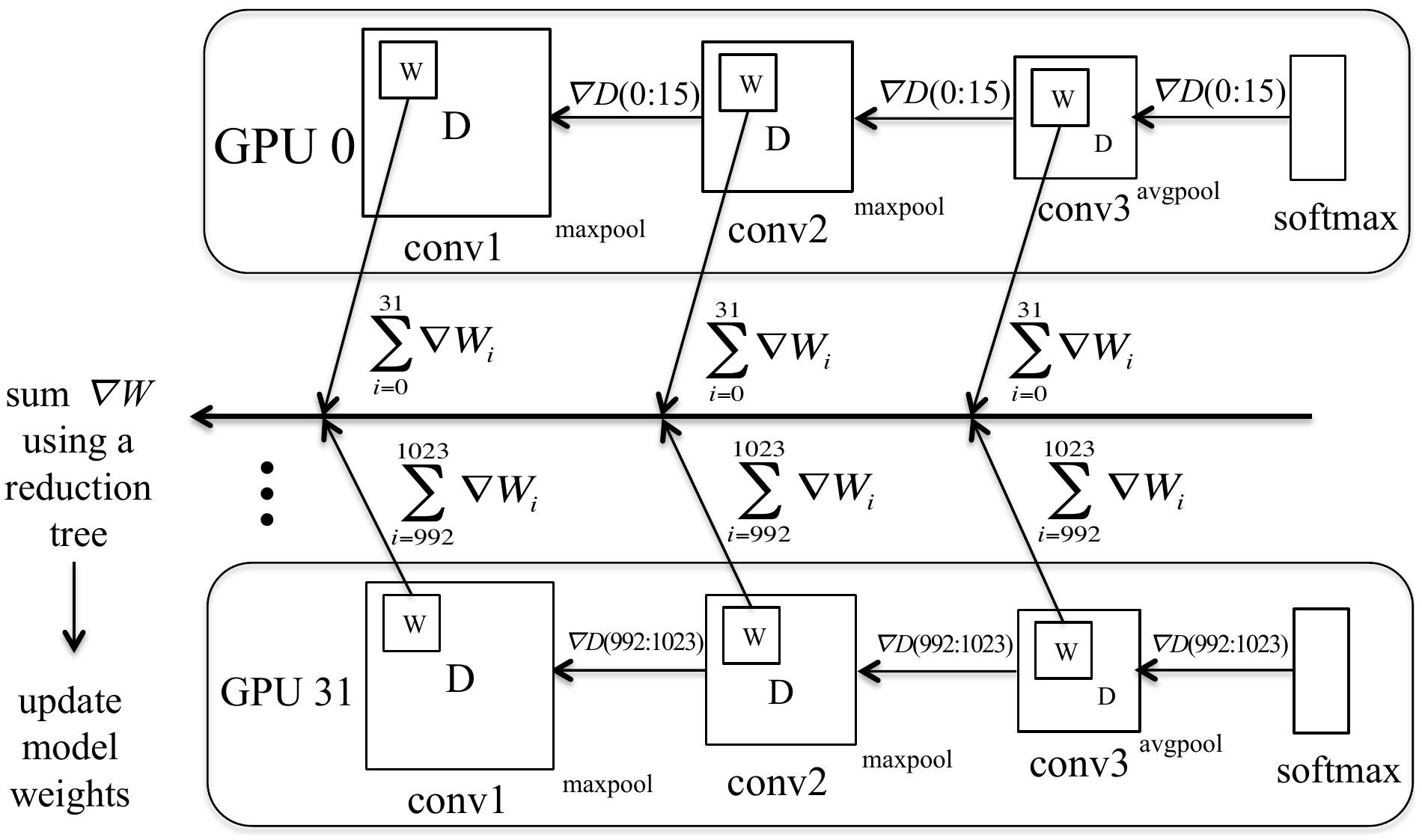}
	\caption{{\bf Data parallel DNN training in FireCaffe:} Each worker (GPU) gets a subset of each batch.}
	\label{fig:data-parallel}
\end{figure}

In computer vision, some of the most popular and accurate DNN models (e.g. GoogLeNet~\cite{googlenet}) consist primarily of convolution layers, where the spatial resolution of the filters is smaller than the resolution of the activations. 
For these convolutional models, data parallelism is typically preferable because it requires less communication -- that is, $|\nabla W|$ is much smaller than $|\nabla D|$ at typical batch sizes.
Notice that the computer vision DNNs in Table~\ref{T:data-volumes} all have this property.
In FireCaffe, we enable data parallelism across a cluster of GPUs, and we find that it produces ample speedups for training popular deep convolutional neural network architectures.
We illustrate our data parallel approach in Figure~\ref{fig:data-parallel}. 
In this configuration, all the GPUs contain the full DNN model parameters. 
Each worker (GPU) gets a subset of each batch. 
The GPUs compute their share of the weight gradients. 
Once the gradients are calculated locally, they are added together using either a parameter server or a reduction tree communication (described in Section \ref{sec:reduction-tree}). 

\begin{table*}[htb]
	\footnotesize
	\caption{Volumes of data and computation for four widely-used DNN architectures. The batch size impacts all numbers in this table except for $|W|$, and we use a batch size of 1024 in this table. Here, TFLOPS is the quantity of computation to perform.}
	\label{T:data-volumes}
	\centering
	
	\begin{tabulary}{17.2cm}{|C|C|C|C|C|C|C|} 
		\hline
		DNN architecture                    & typical use-case  & data\_size $|D|$   & weight\_size $|W|$  & data/weight ratio    & Forward+Backward TFLOPS/batch \\ \hline
		NiN~\cite{NiN}                       & computer vision   & 5800MB               & 30MB                    & 195                      & 6.7TF           \\   \hline
		AlexNet~\cite{Krizhevsky14}  & computer vision     & 1680MB               & 249MB                  & 10.2                     & 7.0TF           \\   \hline
		GoogLeNet~\cite{googlenet}   & computer vision      & 19100MB           & 54MB                    & 358                     & 9.7TF            \\  \hline
		VGG-19~\cite{VGG-19}            & computer vision      & 42700MB           & 575MB                 & 71.7                     & 120TF           \\  \hline
		MSFT-Speech~\cite{Seide11}   & speech recognition  & 74MB                  & 151MB                 & 0.49                     & 0.00015TF           \\  \hline
	\end{tabulary}
\end{table*}

%

\section{Choosing DNN architectures to accelerate} 
\label{sec:choosing-architectures}
\begin{figure}[!t]
	\centering
	\fbox{
		\includegraphics[height=2.5in]{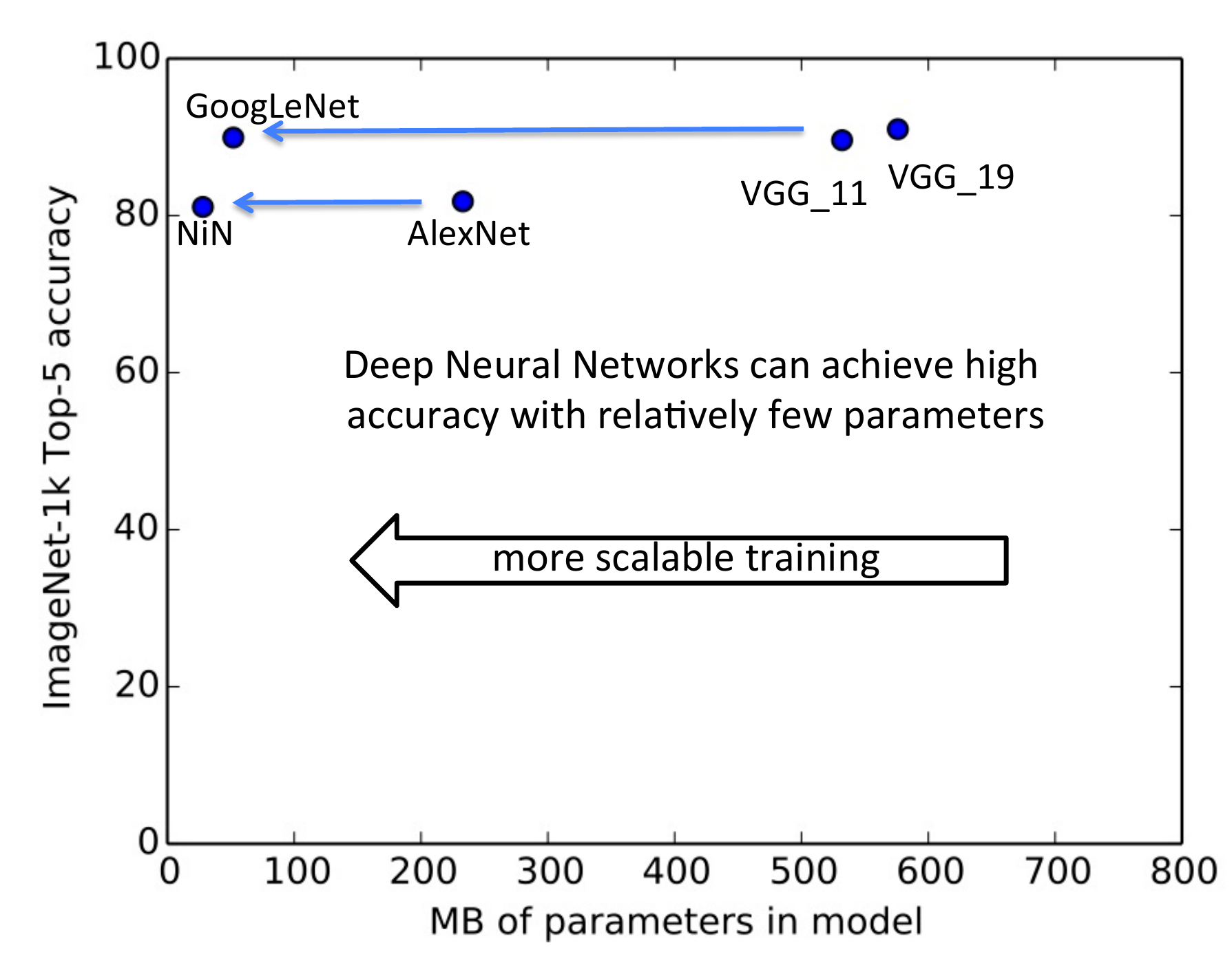}
	}
	\caption{Deep neural network architectures with more parameters do not necessarily deliver higher accuracy.}
	\label{fig:numParams_vs_accuracy}
\end{figure}

\vspace{-0.1in}
Of the popular deep convolutional neural network architectures for computer vision, some are more amenable to data parallel training than others.
One might na\"{\i}vely assume that DNN models with more parameters would produce higher classification accuracy.
To evaluate this assumption, consider Figure~\ref{fig:numParams_vs_accuracy}, where we plot the total size of all parameters in bytes versus top-5 ImageNet accuracy for several popular DNN architectures.
Observe that Nerwork-in-Network (NiN)~\cite{NiN} and AlexNet~\cite{alexnet} have similar accuracy, while NiN has 8x fewer parameters than AlexNet.
Likewise, GoogLeNet~\cite{googlenet} and VGG~\cite{VGG-19} have similar accuracy, yet GoogLeNet has 10x fewer parameters.
In data parallel training, $|\nabla W|$ is the quantity of data sent by each GPU worker, so DNN architectures with fewer parameters require less communication and are more amenable to training at large scale.

You may wonder, what are the architectural choices that led to NiN and GoogLeNet having 8-10x fewer parameters than AlexNet and VGG?
The answer is twofold.
First, GoogLeNet and NiN are more judicious in their use of filters with spatial resolution: many of the filters in GoogLeNet and NiN have a resolution of 1x1 instead of 3x3 or larger.
Second, while VGG and AlexNet each have more than 150MB of fully-connected layer parameters, GoogLeNet has smaller fully-connected layers, and NiN does not have fully-connected layers.

In summary, models with fewer parameters are more amenable to scalability in data parallel training, while still delivering high accuracy.
Therefore, for the rest of the paper, we focus our efforts on accelerating the training of models with fewer parameters (e.g. NiN and GoogLeNet) while maintaining high accuracy.

\section{Implementing efficient data parallel training}
\label{sec:implementation}
\vspace{-0.1in}
Our data-parallel distributed training strategy requires no communication among GPU workers in the forward pass.
In the backward pass, a traditional single-GPU implementation (e.g. single-GPU Caffe~\cite{jia2014caffe}) sums the weight gradients over all images in the batch and then uses the weight gradient sum to update the model.\footnote{However, the data gradients ($\nabla D$) are not summed up.}
When we distribute the backward pass over a compute cluster, each GPU worker computes a sum of the weight gradients ($\sum \nabla W$) for its subset of the batch.
Then, we sum the weight gradients across GPUs.
This gradient aggregation scheme produces identical numerical results as you would find on a single GPU.

Now, our task is to find an efficient way to sum up $\nabla W$ among GPUs in a compute cluster.
We consider two strategies for implementing this gradient aggregation: parameter servers, and reduction trees.

\subsection{Parameter server}
\label{sec:param-server}
\vspace{-0.1in}
One strategy for communicating gradients is to appoint one node as a {\em parameter server}.
The remaining {\em worker} nodes are each assigned a subset of the batch on which to perform forward and backward-propagation.
After each backward pass, all the workers send their gradient updates to the parameter server.
Then, the parameter server computes the sum of the gradients.
Finally, the parameter server sends the summed gradients to the workers, and the workers apply these gradient updates to their local copies of the model.
We illustrate the parameter server communication pattern in Figure~\ref{fig:param_server}.

The logical question here is, {\em what is the communication overhead of a parameter server, and how does that overhead scale as we increase the number of GPU workers?}
Recall from Section~\ref{sec:parallelism-strategies} that each GPU worker provides $|W| = |\nabla W|$ bytes of weight gradients (Equation~\ref{eq:weightsize}), which need to be summed with gradients from all other GPU workers.
Now, the bottleneck is is in sending and receiving all the gradients on one parameter server.
If there are $p$ GPU workers, the parameter server is responsible for sending and receiving $|\nabla W| * p$ bytes of data.
If each node (GPU worker or parameter server) can send and receive data at a rate of $BW$ bytes/s, then we can calculate the minimum communication time as follows:
\begin{small}
\begin{equation} 
\label{eq:param-server}
param\_server\_communication\_time = \frac{|\nabla W| * p}{BW} (sec)
\end{equation}
\end{small}
In other words, the parameter server's communication time scales linearly as we increase the number of GPU workers; doubling the number of workers leads to at least 2x more communication time per gradient update.
We confirm this experimentally in Figure~\ref{fig:param_server_vs_reduction_tree}.

For the parameter server experiments in Figure~\ref{fig:param_server_vs_reduction_tree}, we have implemented a fully synchronous parameter server with the following characteristics.
The parameter server is one arbitrarily-selected server in the cluster, while the other servers are workers; the parameter server and worker servers have identical hardware.
After each batch, the workers send their weight gradients to the parameter server, the parameter server computes the sum, and then the parameter server sends the sum back to the workers.

There are a number of ways to augment the parameter server for greater scalability. 
For example, when having a single parameter server became a bottleneck, Microsoft Adam~\cite{Adam} and Google DistBelief~\cite{DistBelief} each defined a pool of nodes that collectively behave as a parameter server.
The bigger the parameter server hierarchy gets, the more it looks like a reduction tree.
This made us wonder: could we achieve greater scalability if we implement gradient aggregation as a reduction tree?

\subsection{Reduction tree}
\label{sec:reduction-tree}
\vspace{-0.1in}
There are various common patterns of communication in parallel programs; among such common patterns, a frequently occurring one is {\em allreduce}. 
This pattern occurs when each worker produces one or more data values that must be globally reduced (generally with a commutative binary element-wise operator) to produce a single result value, and then this single value must be broadcast to all workers before they can continue. 
In this work, each worker produces a single vector of length $|\nabla W|$ (the gradient updates for that worker), which must be reduced using element-wise vector addition (to sum the per-worker gradient updates for each parameter). 
Since this computation exactly fits the {\em allreduce} communication pattern it is convenient to use existing library support for such operations. 
While there are many possible implementations of {\em allreduce}, most share the key property that the time taken to perform the operation scales as the log of the number of workers (at least for large numbers of workers). 
Intuitively, this is because {\em allreduce} algorithms use binomial reduction tree and/or butterfly communication patterns internally~\cite{MPICHCollectives}.
Out of the possible allreduce implementation strategies, we find that the binomial reduction tree is particularly easy to reason about on a theoretical level.
So, for the rest of this section, we focus on allreduce communication implemented with a reduction tree.

\begin{figure}[!t]
	\centering
	\subfigure[parameter server]{
		\includegraphics[width=2.5in]{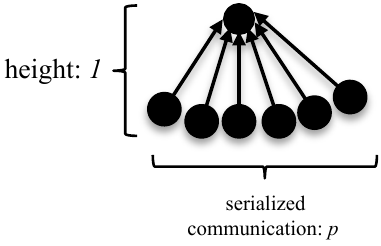}
		\label{fig:param_server}
	}
	\subfigure[reduction tree]{
		\includegraphics[width=2.5in]{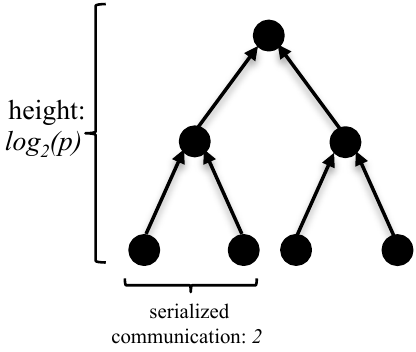}
		\label{fig:reduction_tree}
	}
	\label{fig:diagram_paramserver_reductiontree}
	\caption{Illustrating how parameter servers and reduction trees communicate weight gradients. In this figure, we only show the summing-up of weight gradients. We distribute the weight gradient sums by going back down the tree.}
\end{figure}

\begin{figure}[!t]
	\centering
	\fbox{
		\includegraphics[width=4in]{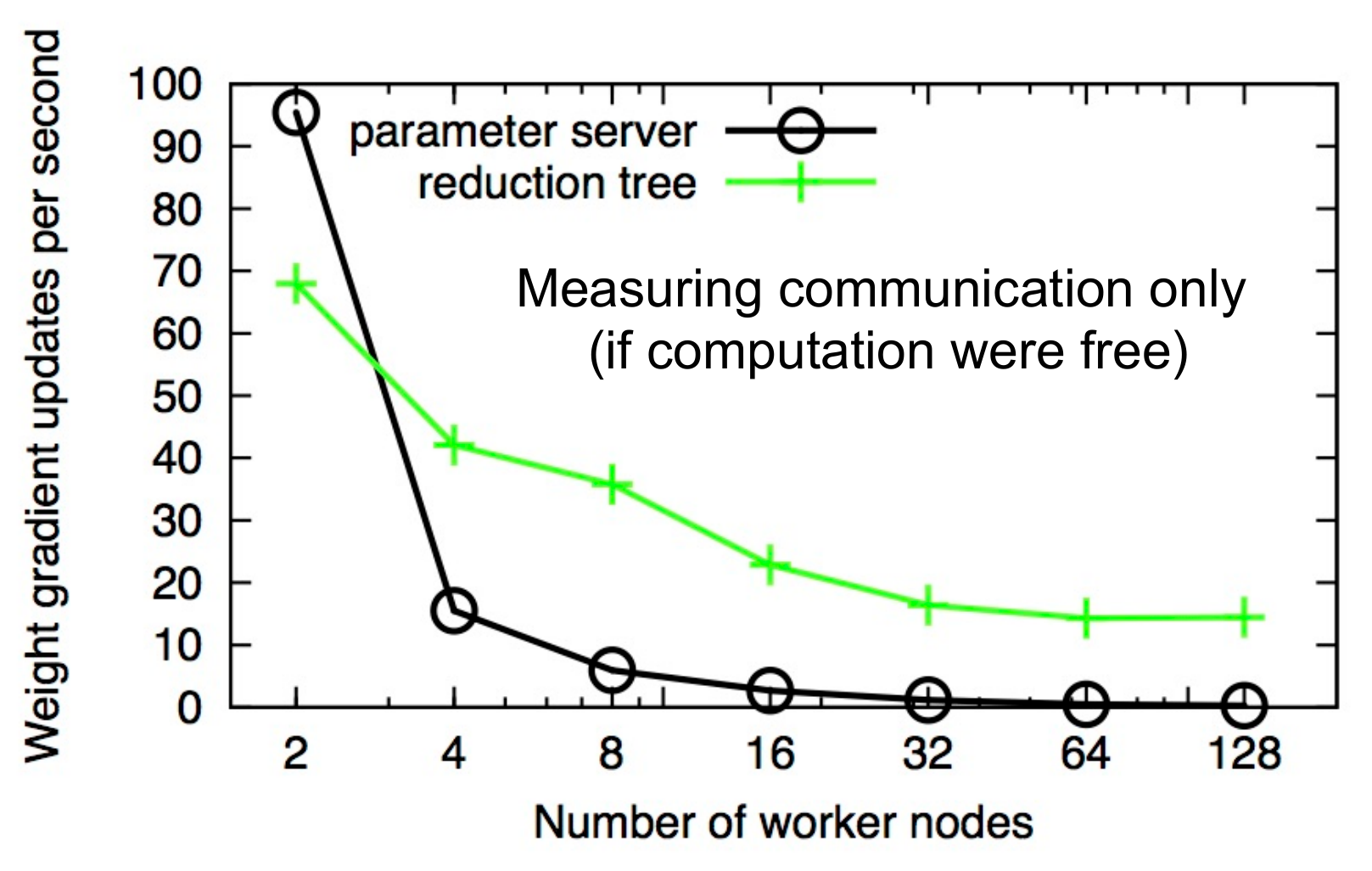}
	}
	\caption{Comparing communication overhead with a parameter server vs. a reduction tree. 
    This is for the Network-in-Network DNN architecture, so each GPU worker contributes 30MB of gradient updates. }
	\label{fig:param_server_vs_reduction_tree}
\end{figure}

In Figures~\ref{fig:param_server} and~\ref{fig:reduction_tree}, we present the intuition on how parameter servers and reduction trees differ. 
We might think of a parameter server as a reduction tree with a height of 1 and a branching factor of $p$.
However, many cluster computers and supercomputers have several dimensions of network fabric among nodes (e.g. an {\em N-D Torus}), which enable nodes to talk to each other via many different paths.
With this in mind, we can sum gradients using a taller reduction tree, where nodes collaboratively sum the gradients.
For example, consider a binary communication tree with a branching factor of 2 and a depth of $log_2(p)$.
In this case, the serialized communication is $2log_2(p)$; the outer $2$ represents the fact that each node receives data from 2 children, and the $log_2(p)$ is the height of the tree. 
Therefore, unlike the parameter server model, the reduction tree's communication time is:
\begin{small}
\begin{equation} 
\label{eq:reduction-tree}
reduction\_tree\_communication\_time = \frac{|\nabla W| * 2log_2(p)}{BW} (sec)
\end{equation}
\end{small}
In practice, the base of $log(p)$ depends on the branching factor in the reduction tree, but the basic idea here is straightforward:
While the parameter server communication overhead scales {\em linearly} with $p$, reduction tree communication is much more efficient because it scales {\em logarithmically} as $O(log(p))$.
We confirm that reduction trees scale more efficiently than parameter servers in Figure~\ref{fig:param_server_vs_reduction_tree}.

\section{Evaluation of FireCaffe-accelerated training on ImageNet}
\label{sec:eval}
\vspace{-0.1in}
In this section, we evaluate how FireCaffe can accelerate DNN training on a cluster of GPUs. 
We train GoogLeNet~\cite{googlenet} and Network-in-Network~\cite{NiN} on up to 128 GPU servers in the Titan supercomputer (described in Section~\ref{sec:hardware}), leveraging FireCaffe's reduction tree data parallelism (Section~\ref{sec:reduction-tree}).
We begin by describing our evaluation methodology, and then we analyze the results.

\subsection{Evaluation Methodology}
\label{sec:eval-method}
\vspace{-0.1in}
We now describe a few practices that aid in comparing advancements in accelerating the training of deep neural networks.\\

\noindent
{\bf 1. Evaluate the speed and accuracy of DNN training on a publicly-available dataset.} \\
In a recent study, Azizpour \etal applied DNNs to more than 10 different visual recognition challenge datasets, including human attribute prediction, fine-grained flower classification, and indoor scene recognition~\cite{Azizpour15}.
The accuracy obtained by Azizpour \etal ranged from 56\% on scene recognition to 91\% on human attribute prediction.
As you can see, the accuracy of DNNs and other machine learning algorithms depends highly on the specifics of the application and dataset to which they are applied.
Thus, when researchers report improvements in training speed or accuracy on proprietary datasets, there is no clear way to compare the improvements with the related literature.
For example, Baidu~\cite{DeepImage} and Amazon~\cite{amazon} recently presented results on accelerating DNN training.
Amazon and Baidu\footnote{Baidu evaluated their training times using proprietary dataset~\cite{DeepImage}. Baidu also did some ImageNet experiments, but Baidu did not report the training time on ImageNet.} each reported their training time numbers on a proprietary dataset, so it's not clear how to compare these results with the related literature. 
In contrast, we conduct our evaluation on a publicly-available dataset, ImageNet-1k~\cite{imagenet}, which contains more than 1 million training images, and each image is labeled as containing 1 of 1000 object categories.
ImageNet-1k is a widely-studied dataset, so we can easily compare our accuracy, training speed, and scalability results with other studies that use this data.\\

\noindent
{\bf 2. Report hyperparameter settings such as weight initialization, momentum, batch size, and learning rate.} \\
Glorot \etal~\cite{xavier}, Breuel~\cite{breuel2015}, and Xu \etal~\cite{prelu_comparison} have each shown that seemingly-subtle hyperparameter settings such as weight initialization can have a big impact on the speed and accuracy produced in DNN training.
When training Network-in-Network (NiN)~\cite{NiN}, we initialize the weights with a gaussian distribution centered at 0, and we set the standard deviation (std) to 0.01 for 1x1 convolution layers, and we use std=0.05 for other layers. 
For NiN, we initialize the bias terms to a constant value of 0, we set the weight decay to 0.0005, and we set momentum to 0.9.
These settings are consistent with the Caffe configuration files released by the NiN authors~\cite{NiN}.

Frustratingly, in Google's technical reports on GoogLeNet~\cite{googlenet, googleBN}, training details such as batch size, momentum, and learning rate are not disclosed.
Fortunately, Wu \etal~\cite{princeton_googlenet} and Guadarrama~\cite{bvlc_googlenet} each reproduced GoogLeNet and released all the details of their training protocols.
As in~\cite{bvlc_googlenet}, we train GoogLeNet with momentum=0.9 and weight decay=0.0002, we use xavier~\cite{xavier} weight initialization, and we initialize the bias terms to a constant value of 0.2.
We will address learning rate and batch size settings in the following sections. 

Given a DNN architecture, there are a number of strategies that can further increase accuracy, albeit at a substantial computational cost.
One such strategy is to train multiple independent copies of a DNN architecture (e.g. GoogLeNet), each with a different random number generator seed for initializing the parameters.
At test time, these DNNs can be used as an {\em ensemble} -- that is, all DNNs are run on the test data, and for each test data item, the DNN's classification activations are averaged. 
For example, using an ensemble of 7 GoogLeNet DNNs, Szegedy \etal achieved a 2 percentage-point accuracy improvement on ImageNet, compared to a single GoogLeNet baseline~\cite{googlenet}.
An other such technique is to augment the data by adding deformations or color variations during training and/or testing~\cite{DeepImage}.
Our focus in this paper is to show speedup on training single models and compare with reported baselines. 
Hence we avoid using exotic data augmentation or ensembles of multiple DNNs. 
In our experiments, we resize images to 256x256; at training time we use a 224x224 crop with a randomized offset, and at test time we classify the 224x224 crop in the center of the image; these settings are also commonly used in the AlexNet~\cite{alexnet} and Network-in-Network~\cite{NiN} DNN architectures.
\\

\noindent
{\bf 3. Measure {\em speedups} with respect to a single-server baseline.} \\
In order to meaningfully measure how much we have accelerated DNN training by adding more GPUs, we must have a representative baseline, e.g. with a single GPU.
When reporting results, we begin by considering time required to train a DNN on single GPU, and we report our multi-GPU speedups with respect to this single-GPU baseline.
A recent study by Microsoft~\cite{Adam} reported training a custom DNN architecture (e.g. not GoogLeNet or NiN) on a cluster of CPU servers.
This may sound impressive, but Microsoft did not report the time that the model would take to train on a single server.
It could be that Microsoft achieved a 10x speedup by going from 1 server to 10 servers, or the speedup could be 2x -- this isn't clear from the information provided in Microsoft's paper. 
This illustrates the importance of measuring the speed of scalable DNN training systems with respect to a single-server baseline. \\

\noindent
{\bf 4. Measure {\em accuracy} with respect to a single-server baseline.} \\
In our experience, if hyperparameters such as learning rate and batch size are selected too aggressively, a DNN model may converge quickly, but fall short of the state-of-art accuracy. 
Therefore, in our experiments, we train multi-GPU models until they reach to the single-GPU accuracy baseline;  this validates that we can accelerate DNN training without degrading accuracy.
However, in cluster-scale multi-GPU training experiments by Baidu~\cite{DeepImage} and Flickr~\cite{yahoo15}, the training is stopped prematurely before the DNNs converge.
This leaves us wondering whether the Baidu and Flickr multi-GPU training experiments would have reproduced the accuracy produced on a single GPU.
To avoid this type of confusion, we evaluate both the speed and accuracy of FireCaffe DNN training with respect to a single-server/single-GPU baseline.

\subsection{Results: Midsized deep models}
\label{sec:midsized}
\vspace{-0.1in}
Using the settings described by Krizhevsky~\cite{alexnet}, we find that AlexNet achieves 58.9\% top-1 ImageNet-1k accuracy after 100 epochs of training. 
After just 47 epochs of training, we find that NiN also converges to 58.9\% top-1 accuracy.
Each training iteration of NiN is more time-consuming than AlexNet, and AlexNet and NiN both take approximately 6 days to converge to this level of accuracy.

At Google, Krizhevsky developed a scheme for accelerating AlexNet training using multiple GPUs within a single server~\cite{Krizhevsky14}.
Krizhevsky's strategy uses data parallelism in convolutional layers and model parallelism in fully-connected layers.
As we show in Table~\ref{T:midsized-models}, Krizhevsky achieves near-linear acceleration up to 8 GPUs, but it has not been shown to scale beyond a single server.
For reasons that we don't entirely understand, Krizhevsky's accuracy drops by 1.8 percentage points when doing multi-GPU training~\cite{Krizhevsky14}.

In FireCaffe, we scale NiN training to 32 GPUs, which is the scale at which we find communication time and computation are approximately equal\footnote{at a batch size of 1024}.
We begin by using the learning rate and batch size settings that were reported in the Caffe configuration file released by the NiN authors~\cite{NiN}: For a batch size of 256, we use an initial learning rate of 0.01, and we reduce this by a factor of 10x twice during our training.
Using this configuration, we reproduce the single-GPU NiN accuracy in 11 hours (13x speedup) when training on 32 GPUs.

For a fixed number of epochs, increasing the batch size reduces the number of times we need to communicate weight gradients, thus reducing the overall training time.
With this in mind, we now train NiN with a batch size of 1024.\footnote{While keeping a fixed number of epochs. In other words, with a batch size of 1024, we perform 4x fewer training iterations than with a batch size of 256.}
As in~\cite{Krizhevsky14} when we increase the batch size, we increase the learning rate by an equal proportion.
For example, when we use a batch size of 1024, we initialize the learning rate to 0.04.
In this configuration, we train NiN in just 6 hours (23x speedup) on 32 GPUs.
By increasing the batch size to 1024, we achieved a substantial speedup, but this came at the price of reducing the final accuracy by $\frac{3}{10}$ of a percentage point.
We expect that this $\frac{3}{10}$\% of accuracy could be regained at a batch size of 1024 -- while retaining a substantial speed advantage -- by training for a few more epochs.
Finally, on 128 GPUs, we achieve a 39x speedup over single-GPU training.

So far, we have compared FireCaffe to the cuda-convnet2 framework from Google~\cite{Krizhevsky14}, which runs on a single-server/multi-GPU platform but not in a multi-server distributed platform.
In addition to cuda-convnet2, Google has developed the TensorFlow framework~\cite{TensorFlow}, which also supports single-server/multi-GPU training but not distributed multi-server training. 
Thus far, Google has not released training speed results for multi-GPU TensorFlow. 
Twitter~\cite{TwitterDNN} has also experimented with scaling DNN training to 8 GPUs, but speed and accuracy results have not been released.
Tencent~\cite{tencent}, Theano~\cite{theanoMultiGPU}, and Facebook~\cite{fbMultiGPU} have published AlexNet single-server/multi-GPU training times that are slower than Google~\cite{Krizhevsky14}.
Other than FireCaffe, we have not seen literature on training AlexNet/NiN-scale models in a multi-server/multi-GPU setting.
On 32 GPUs, FireCaffe is at least 3x faster to train AlexNet/NiN-scale models than all of the aforementioned results.

\begin{table*}[htb]
	\footnotesize
	\caption{Accelerating the training of midsized deep models on ImageNet-1k.}
	\label{T:midsized-models}
	\centering
	\begin{tabulary}{17cm}{|p{1in}|p{1.2in}|C|C|C|C|C|C|C|} 
		\hline
		& Hardware                                          & Net                             & Epochs  & Batch size & Initial Learning Rate & Train time   & Speedup~   & Top-1 Accuracy       \\ \hline
		Caffe~\cite{jia2014caffe}         & 1 NVIDIA K20                                   & AlexNet \cite{alexnet} & 100        & 256           & 0.01                         &  6.0 days    & 1x             & 58.9\%  \\ \hline
		Caffe                                       & 1 NVIDIA K20                                   & NiN \cite{NiN}            & 47          & 256          & 0.01                          & 5.8 days    & 1x              & 58.9\%  \\ \hline
		Google cuda-convnet2 ~\cite{Krizhevsky14}    & 8 NVIDIA K20s (1 node)                    & AlexNet                       & 100       & varies       & 0.02                          & 16 hours   & 7.7x           &  57.1\%  \\ \hline
		FireCaffe (ours)                       & 32 NVIDIA K20s (Titan supercomputer)   & NiN                       & 47         & 256          & 0.01                          & 11 hours    & 13x           & 58.9\%  \\ \hline
		FireCaffe-batch1024 (ours)     & 32 NVIDIA K20s (Titan supercomputer)   & NiN                       & 47         & 1024        & 0.04                           & 6 hours     &  23x    & 58.6\%  \\ \hline
		FireCaffe-batch1024 (ours)     & 128 NVIDIA K20s (Titan supercomputer)   & NiN                       & 47         & 1024        & 0.04                           & 3.6 hours     & {\bf 39x}    & 58.6\%  \\ \hline
	\end{tabulary}
\end{table*}

\subsection{Results: Ultra-deep models}
\label{sec:ultra-deep}
\vspace{-0.1in}
Ultra-deep models such as GoogLeNet can produce higher accuracy, but they present an even bigger challenge in terms of training time.
Internally, Google has trained GoogLeNet on a cluster of CPU servers, but they have not reported the time required to complete this training~\cite{googlenet, googleBN}.
Fortunately, Guadarrama reproduced GoogLeNet in Caffe, and he released his GoogLeNet Caffe configuration files~\cite{bvlc_googlenet}.
Guadarrama trained for 64 epochs using a batch size of 32 and an initial learning rate of 0.01, and we use these settings in our single-GPU GoogLeNet training experiments.
Instead of occasionally reducing the learning rate by 10x, Guadarrama used a polynomial learning rate -- that is, the learning rate is gradually reduced after every iteration of training. 
More specifically, at a given iteration of training, the learning rate is calculated as $initialLearningRate(1 - \frac{iter}{max\_iter})^{power}$, and we set $power$ to 0.5 in all of our GoogLeNet training runs.
Running this in Caffe on a single-GPU, GoogLeNet takes 21 days to train on ImageNet-1k, producing 68.3\% top-1 accuracy and 88.7\% top-5 accuracy.
This is slightly lower than the 89.9\% top-5 single-model accuracy reported by Google~\cite{googlenet}, and it will be interesting to see whether the open-source Caffe community will eventually be able reproduce or surpass Google's GoogLeNet accuracy.
Here, we use the single-GPU Caffe GoogLeNet accuracy (88.7\% top-5 accuracy) as a baseline, and we aim to reproduce this rapidly on a cluster of GPUs. 

Now, we consider how to accelerate GoogLeNet training using FireCaffe.
We initially tried to run GoogLeNet with a batch size of 32 on a GPU cluster, but there just wasn't enough work per batch to keep a GPU cluster saturated.
As we learned earlier in the paper, larger batch sizes lead to less frequent communication and therefore enable more scalability in a distributed setting.
When modifying the batch size, Breuel~\cite{breuel2015} and Krizhevsky~\cite{Krizhevsky14} found that the choice of learning rate is crucial in order to preserve high accuracy.
We trained five separate versions of GoogLeNet, each with a different initial learning rate (LR): \{0.02, 0.04, 0.08, 0.16, and 0.32\}, and all with a batch size of 1024.
With LR=0.16 and LR=0.32, GoogLeNet didn't ever learn anything beyond random-chance accuracy on the test set.
Using LR=0.02 produced 66.1\% top-1 ImageNet-1k accuracy, and LR=0.04 produced 67.2\%.
Finally, we declare victory with LR=0.08, where we achieved 68.3\% accuracy (again, with a batch size of 1024), which matches the accuracy of the baseline that used a batch size of 32.
With a batch size of 1024 and a fixed number of epochs, we find that FireCaffe on 32 GPUs can train GoogLeNet 23x faster than a single GPU.
When we move from a batch size of 32 with LR=0.01 to a batch size of 1024 with LR=0.08, we find that GoogLeNet takes a few more epochs to converge (72 epochs instead of 64 epochs), so the absolute training speedup is 20x; we show these results in Table~\ref{T:enormous-models}.
In other words, FireCaffe can train GoogLeNet in 23.4 hours on 32 GPUs, compared to 21 days on a single GPU.
Finally, on 128 GPUs, we achieve a 47x speedup over single-GPU GoogLeNet training, while matching the single-GPU accuracy.

\begin{table*}[htb]
	\footnotesize
	\caption{Accelerating the training of ultra-deep, computationally intensive models on ImageNet-1k.}
	\label{T:enormous-models}
	\centering
	\begin{tabulary}{17.2cm}{|p{0.5in}|p{1.2in}|C|C|C|C|C|C|C|C|} 
		\hline
		& Hardware                                                 & Net                                         & Epochs  & Batch size  & Initial Learning Rate     & Train time   &  Speedup~ & Top-1 Accuracy   & Top-5 Accuracy   \\ \hline
		Caffe                    & 1 NVIDIA K20                                         & GoogLeNet \cite{googlenet}  & 64          & 32                & 0.01  & 21 days       & 1x             & 68.3\%                & 88.7\%  \\ \hline 
		FireCaffe (ours)    & 32 NVIDIA K20s (Titan supercomputer)  & GoogLeNet                           & 72          & 1024            & 0.08  & 23.4 hours      & 20x    & 68.3\%               & 88.7\% \\ \hline 
		FireCaffe (ours)    & 128 NVIDIA K20s (Titan supercomputer)  & GoogLeNet                           & 72          & 1024            & 0.08  & 10.5 hours      & {\bf 47x}    & 68.3\%               & 88.7\% \\ \hline 
	\end{tabulary}
\end{table*}

\section{Complementary approaches to accelerate DNN training}
\label{sec:related}
\vspace{-0.1in}
We have discussed related work throughout the paper, but we now provide a brief survey of additional techniques to accelerate deep neural network training.
Several of the following techniques could be used in concert with FireCaffe to further accelerate DNN training.

\subsection{Accelerating convolution on GPUs}
\vspace{-0.1in}
In the DNN architectures discussed in this paper, more than 90\% of the floating-point operations in forward and backward propagation reside in convolution layers, so accelerating convolution is key to getting the most out of each GPU. 
Recently, a number of techniques have been developed to accelerate convolution on GPUs.
Unlike CPUs, NVIDIA GPUs have an inverted memory hierarchy, where the register file is larger than the L1 cache.
Volkov and Demmel~\cite{Volkov:08} pioneered a {\em communication-avoiding} strategy to accelerate matrix multiplication on GPUs by staging as much data as possible in registers while maximizing data reuse.
Iandola \etal~\cite{IandolaConv13} extended the communication-avoiding techniques to accelerate 2D convolution; and cuDNN~\cite{cuDNN} and maxDNN~\cite{maxDNN} extended the techniques to accelerate 3D convolution.
FireCaffe can be coupled with current and future GPU hardware and convolution libraries for further speedups. 

\subsection{Reducing communication among servers}
\vspace{-0.1in}

Reducing the quantity of data communicated per batch is a useful way to increase the speed and scalability of DNN training.
There is an inherent tradeoff here: as gradients are more aggressively quantized, training speed goes up, but the model's accuracy may go down compared to a non-quantized baseline.
While FireCaffe uses 32-bit floating-point values for weight gradients, Jeffrey Dean stated in a recent keynote speech that Google often uses 16-bit floating-point values for communication between servers in DNN training~\cite{DeanGTC}.
Along the same lines, Wawrzynek \etal used 16-bit weights and 8-bit activations in distributed neural network training~\cite{SpertII}.
Going one step further, Seide \etal used 1-bit gradients for backpropagation, albeit with a drop in the accuracy of the trained model~\cite{1bit}.
Finally, a related strategy to reduce communication between servers is to discard (and not communicate) gradients whose numerical values fall below a certain threshold. 
Amazon presented such a thresholding strategy in a recent paper on scaling up DNN training for speech recognition~\cite{amazon}.
However, Amazon's evaluation uses a proprietary dataset, so it is not clear how this type of thresholding impacts the accuracy compared to a well-understood baseline.

So far in this section, we have discussed strategies for compressing or quantizing data to communicate in distributed DNN training.
There has also been a series of studies on applying dimensionality reduction to DNNs once they have been trained.
Jaderberg \etal~\cite{jaderberg2014} and Zhang \etal~\cite{zhang2014} both use PCA to compress the weights of DNN models by up to 5x, albeit with a substantial reduction in the model's classification accuracy.
Han \etal~\cite{han2015} use a combination of pruning, quantization, and Huffman encoding to compress the weights of pretrained models by 35x with no reduction in accuracy.
Thus far, these algorithms have only been able to accelerate DNNs at test time.

\section{Conclusions}
\label{sec:conclusions}
\vspace{-0.1in}
Long training times impose a severe limitation on progress in deep neural network research and productization.
Accelerating DNN training has several benefits.
First, faster DNN training enables models to be trained on ever-increasing dataset sizes in a tractable amount of time.
Accelerating DNN training also enables product teams to bring DNN-based products to market more rapidly.
Finally, there are a number of compelling use-cases for real-time DNN training, such as robot self-learning. 
These and other compelling applications led us to focus on the problem of accelerating DNN training, and our work has culminated in the FireCaffe distributed DNN training system.

Our approach to accelerating DNN training at scale has three key pillars.
First, we select network hardware that achieves high bandwidth between GPU servers -- Infiniband or Cray interconnects are ideal for this.
Second, when selecting a communication algorithm, we find that reduction trees are more efficient and scalable than the traditional parameter server approach. 
Third, we optionally increase the batch size to reduce the total quantity of communication during DNN training, and we identify hyperparameters that allow us to reproduce the small-batch accuracy while training with large batch sizes.
These three pillars helped us to achieve a near-linear speedup for a number of leading deep neural network architectures.
In particular, we have achieved 39x speedup on NiN training, and a 47x speedup on GoogLeNet training on a 128 GPU cluster.


\section*{Acknowledgements}
This research used resources of the Oak Ridge Leadership Facility at the Oak Ridge National Laboratory, which is supported by the Office of Science of the U.S. Department of Energy under Contract No. DE-AC05-00OR22725.
Thanks to Bryan Catanzaro, Evan Shelhamer, and Trevor Darrell for helpful discussions on scalable deep neural network training.
Thanks to Sergio Guadarrama and Christian Szegedy for helpful discussions about GoogLeNet.
Thanks to Thomas Breuel for sharing his intuition with us about the design space of deep neural network architectures, and in particular for discussions regarding the impact of learning rates.
Thanks to Kostadin Ilov for IT support and for answering our numerous questions about hardware infrastructure.
Thanks to Aditya Devarakonda for helpful discussions about supercomputer network interconnect hardware.

{\small
\bibliographystyle{ieee}
\bibliography{bibliography}
}

\section*{Appendix: Frequently Asked Questions about FireCaffe}
\noindent
1. {\em I use Caffe. Will my DNN models and network definitions work with FireCaffe?} \\ 
Yes. \\

\noindent
2. {\em What are some good DNN architectures to use with FireCaffe?} \\
We offer intuition and analysis of this in Sections~\ref{sec:parallelism-strategies} and~\ref{sec:implementation}, but here we offer a short summary of what you need to know.
As a general rule, DNN architectures with fewer parameters train faster in FireCaffe. 
We have found that NiN~\cite{NiN} and AlexNet~\cite{alexnet} provide similar accuracy on ImageNet and other popular data benchmarks.
However, NiN has 8x fewer parameters and therefore incurs 8x less communication cost than AlexNet.
If you were planning to use AlexNet, we recommend trying NiN for faster training. \\

\noindent
3. {\em I want to design my own DNN architectures that will scale well nicely in FireCaffe while producing good accuracy on my problem. What design tradeoffs should I consider?} \\
If you're designing your own DNN architectures, do your best to economize on parameters. 
Also, reducing the convolution and pooling strides may improve accuracy; this doesn't require additional parameters and doesn't hurt training scalability. \\ 

\noindent
4. {\em Is it better to use FireCaffe with one server that contains many GPUs, or is it better to distribute the GPUs distributed over many servers?} \\
FireCaffe is compatible with both of these scenarios. 
FireCaffe can even run across many servers that each contain many GPUs.  
To avoid stragglers during backpropagation, we prefer running FireCaffe on a collection of identical GPUs (e.g. all Titan X or all K80; but preferrably not a mixture of K80 and Titan X).
For optimal speed and utilization, FireCaffe prefers a low-latency interconnect such as Infiniband between servers. \\

\noindent
5. {\em How does FireCaffe handle the storage and loading of training data?} \\
As in Caffe, FireCaffe can ingest LMDB databases~\cite{LMDB} of training data.
This data format is agnostic to the type of data (e.g. images, audio, text), so long as each training data item consists of a vector (e.g. pixels, audio waveform, text trigram) and a label (e.g. dog or cat).
We store the LMDB database of training data on a distributed filesystem that is accessable to all workers.
During DNN training, each worker is responsible for loading its own training data from the distributed filesystem. \\

\noindent
6. {\em Does FireCaffe use nondeterministic techniques such as Hogwild~\cite{hogwild}?} \\
No. 
Given a specific seed for the random number generator, FireCaffe produces repeatable numerical results, just like ordinary Caffe. 
In fact, for a model with Dropout disabled, FireCaffe will give you the exact same numerics as ordinary Caffe.
With Dropout enabled, FireCaffe handles randomization in a slightly different way than Caffe, but the numerics are still deterministic for FireCaffe with a constant number of GPUs. \\

\noindent
7. {\em How did FireCaffe get its name?} \\
Asanovic and Patterson presented a roadmap for warehouse-scale computing in the year 2020, which they call {\em FireBox}~\cite{FireBox}. 
The grand vision is that a typical large-scale commercial datacenter will have extremely low latency connections within and across racks, likely using photonic networking hardware.
Low-latency network hardware is crucial not only for today's mainstream applications like distributed databases and search engines, but also for emerging applications such as deep neural network training at large scale.
We have designed FireCaffe with FireBox-style warehouse-scale computing in mind.

\end{document}